\documentclass[aps,pre,twocolumn,reprint,superscriptaddress,showkeys]{revtex4-2}

\usepackage{amsmath,amssymb,graphicx,bbold}
\usepackage{graphicx}  
\usepackage{dcolumn}  
\usepackage{bm}

\usepackage[caption=false,position=top,margin=12pt,justification=raggedright,singlelinecheck=false,subrefformat=parens,labelformat=parens]{subfig}
\usepackage{subfig}
\usepackage{amsmath}    
\usepackage{graphics}
\usepackage{verbatim}  
\usepackage{color}    
\usepackage{url}
\usepackage{breakurl}
\usepackage[breaklinks]{hyperref}  
\hypersetup{colorlinks, citecolor=blue, linkcolor=blue, urlcolor=blue}
\usepackage{lineno} 
\usepackage{textcomp}
\usepackage{cleveref}
\usepackage{enumitem}

\usepackage[normalem]{ulem}

\begin{document}

\title{Symmetry-Aware Reservoir Computing}

\author{Wendson A. S. Barbosa}
\email{desabarbosa.1@osu.edu}
\affiliation{Department of Physics, Ohio State University, 191 W. Woodruff Ave., Columbus, OH 43210, USA}

\author{Aaron Griffith}
\affiliation{Department of Physics, Ohio State University, 191 W. Woodruff Ave., Columbus, OH 43210, USA}

\author{Graham E. Rowlands}
\email{graham.rowlands@raytheon.com}
\affiliation{Quantum Engineering and Computing, Raytheon BBN Technologies, Cambridge, MA 02138, USA}

\author{Luke C. G. Govia}
\affiliation{Quantum Engineering and Computing, Raytheon BBN Technologies, Cambridge, MA 02138, USA}

\author{Guilhem J. Ribeill}
\affiliation{Quantum Engineering and Computing, Raytheon BBN Technologies, Cambridge, MA 02138, USA}

\author{Minh-Hai Nguyen}
\affiliation{Quantum Engineering and Computing, Raytheon BBN Technologies, Cambridge, MA 02138, USA}

\author{Thomas A. Ohki}
\affiliation{Quantum Engineering and Computing, Raytheon BBN Technologies, Cambridge, MA 02138, USA}

\author{Daniel J. Gauthier}
\email{gauthier.51@osu.edu}
\affiliation{Department of Physics, Ohio State University, 191 W. Woodruff Ave., Columbus, OH 43210, USA} 

\begin{abstract} 

We demonstrate that matching the symmetry properties of a reservoir computer (RC) to the data being processed dramatically increases its processing power. We apply our method to the parity task, a challenging benchmark problem that highlights inversion and permutation symmetries, and to a chaotic system inference task that presents an inversion symmetry rule. 
For the parity task, our symmetry-aware RC obtains zero error using an exponentially reduced neural network and training data, greatly speeding up the time to result and outperforming artificial neural networks.
When both symmetries are respected, we find that the network size $N$ necessary to obtain zero error for 50 different RC instances scales linearly with the parity-order $n$. Moreover, some symmetry-aware RC instances perform a zero error classification with only $N=1$ for $n\leq7$.   
Furthermore, we show that a symmetry-aware RC only needs a training data set with size on the order of $(n+n/2)$ to obtain such performance, an exponential reduction in comparison to a regular RC which requires a training data set with size on the order of $n2^n$ to contain all $2^n$ possible $n-$bit-long sequences. 
For the inference task, we show that a symmetry-aware RC presents a normalized root-mean-square error three orders-of-magnitude smaller than regular RCs. %
For both tasks, our RC approach respects the symmetries by adjusting only the input and the output layers, and not by problem-based modifications to the neural network.  
We anticipate that generalizations of our procedure can be applied in information processing for problems with known symmetries.

\end{abstract}

\maketitle 


\section{Introduction}

Reservoir computing \cite{Jaeger2004,Maass2002,Gauthier2018} is an emerging machine learning (ML) paradigm based on artificial neural networks (ANNs) that is ideally suited for a variety of tasks such as learning dynamical systems from time series data \cite{Pathak2018,Klos2020} or classifying structures in data \cite{Jalalvand2015,Shani2019}.  In comparison to other ML approaches, reservoir computing requires much smaller data sets for training and the training time can be orders-of-magnitude faster while maintaining high performance \cite{vlachas2019,Chattopadhyay2019}, making them suitable for deployment on edge-computing devices \cite{Canaday2018}. 

\begin{figure}[b]
\centering
	\subfloat{%
	\includegraphics[width=\linewidth]{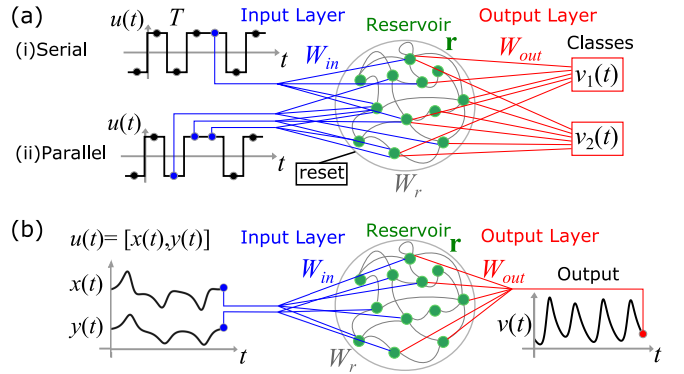}%
	\label{fig:RC_scheme_a}%
	}
	\begin{minipage}[t]{0\textwidth}
	\subfloat{%
	\label{fig:RC_scheme_b}%
	}
    \end{minipage}
\caption{Reservoir Computer scheme for (a) parity task and (b) Lorenz '63 system inference task. }
\label{fig:RC_scheme}
\end{figure}

The core of an RC is a pool of $N$ artificial neurons with recurrent connections, known as the reservoir and illustrated in Fig.~\ref{fig:RC_scheme}, along with an input layer that broadcasts the input data to the reservoir and an output layer that forms a weighted sum of the values of the reservoir nodes that provides the computation result.  Differing from other approaches, the relative weights of the connections of the input layer $W_{in}$ and within the reservoir $W_r$ are generated randomly at instantiation of the RC and held fixed, although their overall scale can be adjusted.  Only the weights of the output layer $W_{out}$ are adjusted during training, which is a linear optimization problem that can be solved using standard tools and is the cause of the short training time.

Even though the RC is a complex network with random weights, it still possesses symmetries that can substantially impact the RC performance depending on the symmetries of the data being processed.  This point was noted and addressed in an \textit{ad hoc} way when using an RC to forecast the dynamics of the Lorenz '63 chaotic attractor \cite{Pathak2017,Lu2017,Lu2018,Griffith2019} and the multi-scale Lorenz `96 system \cite{Chattopadhyay2019}. Failures in such predictions are due to inversion symmetries in both RC and the learning system and can be solved by breaking the RC symmetry \cite{Herteux2020}. 
Symmetry has also been shown to be important when addressed in other ML approaches like deep learning, {\it{e.g.}}, by considering permutation invariant functions to create deep networks that can operate on sets with possibly different sizes \cite{Zaheer2018,Murphy2019} or by adding special layers to feed-forward neural networks to embed physical symmetries \cite{Mattheakis2020}.

Here, we demonstrate for two different tasks that matching the RC and the learning system symmetries by only making straightforward changes to the RC input and output without changing the reservoir can increase the RC performance. To illustrate symmetry matching RCs, we study a classification and an inference task that especially highlight the issue of the symmetry differences between the data and the RC. 

For the classification task, the RC computes the parity of a sequence of digital bits, which is a known challenging ML task because the problem is linearly inseparable \cite{Thornton1996,Grochowski2009,Shalev-Shwartz2017}. Hand-crafted ANNs can tackle this problem with different scaling rules for the number of nodes (see, for example, Refs. \cite{Hertz1991,Wilamowski2003,Hunter2012,Arslanov2016}), but generic ANNs require that the network size \cite{Minsky1969} and training time \cite{Grochowski2009} increase exponentially with the parity order $n$ (defined precisely below) to reach a user-defined accuracy.  We show that the `symmetry-aware' RC requires exponentially smaller $N$ and training data in comparison to the non-aware RC, and has similar or better performance than the hand-crafted ANNs.  

The second task we address is inferring one unknown variable of the Lorenz '63 chaotic dynamical system \cite{Lorenz63} having knowledge of the others. For this task, our RC reduces the normalized root-mean-square error (NRMSE) by three orders-of-magnitude in comparison to a traditional RC.  Furthermore, we demonstrate how to realize such an RC, whose hyperparameters can be discovered automatically using optimization tools \cite{Yperman2016,Griffith2019}. This work paves the way for improving the performance of RCs on other tasks matching the RC and the known symmetries of the learning system by adjusting symmetry-breaking parameters accordingly.

The rest of the paper is organized as follows. In Secs. \ref{sec:parity} and \ref{sec:inference}, we formally introduce 
the parity task and the Lorenz '63 inference task, respectively. We describe the parity-order and the sequence-order permutation symmetries of the parity function and the inversion symmetry present in the Lorenz '63 system. In Sec. \ref{sec:RC}, we introduce the theoretical background of a general RC, followed by brief descriptions of the training procedure and the RC hyperparameters. Section \ref{sec:SymmetryRC} is dedicated to the explanations of the symmetry properties of a regular RC and how it can be modified to match previously known symmetries of the learning system, thus creating a symmetry-aware RC. Finally, in Sec. \ref{sec:Results} we discuss the performance of the symmetry-aware RC and compare it to standard RC results for both the parity and the inference tasks before present our conclusions in Sec. \ref{sec:Conclusion}. 


\section{The Parity Task}
\label{sec:parity}

The task we first consider is to determine the parity of each sequence of $n$ bits in a signal $u(t)$, which is a Boolean time series where each bit has a time duration $T$ and assumes either value +1 or -1. The RC is trained to predict the $n^{th}$ order parity function
\begin{equation}
P_n(t)=\prod_{i=0}^{n}u(t-i T). 
\label{eq:parity}
\end{equation}

\noindent Inspection of this expression reveals two symmetries:
\noindent
\begin{itemize}	[labelsep=*,leftmargin=1pc]
\item \textbf{Parity-order symmetry:} The parity function has an inversion symmetry that depends on $n$. For $n$ odd, an $n$-bit sequence will have the parity changed from $p$ to $-p$ if all its bits are flipped, \textit{i.e.}, $(u,p)\rightarrow (-u,-p)$. On the other hand, $(u,p)\rightarrow (-u,p)$ for $n$ even.
\item \textbf{Sequence-order permutation symmetry:} The parity of a sequence is the same under permutation of its bits. Thus, the parity only depends on the number of positive (or negative) bits in the sequence. 
\end{itemize}

For future reference, we divide the $2^n$ possible $n$-bit input sequences into sets $L_n(l)$ of size $\binom{n}{l}$ according to the number of ones $l$ in the sequence. For each $n$, there are $n + 1$ such sets. Because all $n$-bit sequences containing $l$ ones are equivalent under the permutation symmetry and consequently have the same parity, it should be possible to train a symmetry-aware RC that shares this symmetry with a small number of sequences that cover these $n + 1$ distinct sets, rather than all $2^n$ possible inputs.


\section{The inference task}
\label{sec:inference}
This task is to infer an inaccessible variable of a dynamical system having knowledge of the others. We consider the Lorenz '63 chaotic system and assume that all three variables $x$, $y$ and $z$ are accessible for a training time interval. The RC is trained to infer $z$ having $u=[x,y]$ as input. After the training phase, we only have access to $x$ and $y$. The Lorenz '63 chaotic system with the standard parameters \cite{Lorenz63} is described by
\begin{align}
\dot{x}&=10(y-x)  \nonumber \\ 
 \dot{y}&=x(28-z)-y \\ 
 \dot{z}&=xy-\frac{8}{3}z.\nonumber 
\label{eq:Lorenz}
\end{align} 

\noindent These equations possess an inversion symmetry  $(x,y,z)\rightarrow (-x,-y,z)$, {\it{i.e.}} , for the inference task of $z$, both inputs $u=[x,y]$ and $-u=[-x,-y]$ lead to the same output $z$. This symmetry is similar to the parity-order symmetry for even $n$.


\section{The RC}
\label{sec:RC}
In our RC implementation, also known as an echo state network, the reservoir nodes dynamics $\mathbf{r}$ is governed by    
\begin{equation}
\dot{\bf{r}}(t)=-\gamma {\bf{r}}(t) + \gamma f(W_r{\bf{r}}(t)+W_{in}{\bf{u}}(t) + b) ,
\label{eq:reservoir}
\end{equation}
where $\gamma$ is the decay rate, $f(\cdot)$ is the nonlinear activation function, and $b$ is a bias. While $\gamma$ and $b$ can be different for each node, we take them the same for simplicity. 
While our reservoir is continuous in time and governed by an ordinary differential equation as in Refs. \cite{Gauthier2018, Lu2018,Griffith2019}, other works use a discrete time version of the reservoir such as in Refs.  \cite{Pathak2018, Pathak2017, Lu2017}, for example. Performing a forward Euler integration on Eq. \ref{eq:reservoir} and rescaling $\gamma$ by the integration step recovers the discrete time model. Thus, the two approaches are equivalent and the results presented in the following sections should hold equally well for both approaches.  

The reservoir output is given by 
\begin{equation}
{\bf{v}}(t)=W_{out}g({\bf{r}}(t)), 
\label{eq:readout}
\end{equation}

\noindent where $g(\cdot)$ is often taken as a linear function but we allow it to be nonlinear in order to adjust the RC symmetry as described below. Here, ${\bf{v}}(t) \widehat{=}  z(t)$ is a scalar for the Lorenz '63 inference task, while it is a two-component vector ${\bf{v}}(t) = \{v_1(t),v_2(t)\}$ for the parity task, where it projects the reservoir states onto the parity labels as shown in Fig.~\ref{fig:RC_scheme}. The final RC output parity is $+1$ for each time span $T$ if the average over $\Delta T$ component $\overline v_1$ is larger than $\overline v_2$, and $-1$ otherwise. Here, $\Delta T$ is the measurement window within $T$ used for the reservoir output calculation, which starts at an initial time $T_0$ and finishes at $T_0 + \Delta T$. 

Training the RC uses supervised learning, where an input drives the reservoir and the desired output $Y$ is previously known. We use Ridge regression to find  the output matrix $W_{out}$  by minimizing 
\begin{equation}
|Y - W_{out}g({\bf{r}})|^ 2 + \alpha|| W_{out}||^2,  
\label{eq:Ridge}
\end{equation}

\noindent where the Ridge parameter $\alpha$ prevents overfitting.

The RC is instantiated by choosing randomly the components of $W_{in}$  from a zero-mean normal distribution with variance $\rho_{in}$ and probability $\sigma$ for a non-zero coefficient that specifies the input connectivity. The adjacency matrix $W_r$ has a spectral radius $\rho_r$ and each node has $k$ connections from other reservoir nodes. The hyperparameters $\gamma,\rho_r,\sigma,\text { and } \rho_{in}$  (also  $T_0$ and $\Delta T$ for the parity task) are selected using a Bayesian optimizer \cite{Yperman2016,Griffith2019} (see Appendix \ref{appendix:a}).


\section{A symmetry-aware RC}
\label{sec:SymmetryRC}
First, we describe how a standard RC does not take advantage of the symmetries described above. In previous works that solve the parity task with RC \cite{Bertschinger2004,Dasgupta2012,Snyder2013,Schumacher2013,Coulombe2017,Dion2018,Furuta2018,Kanao2019,Tsunegi2019,Watt2020}, $\mathbf{u}$ is injected into the reservoir as serial data, as shown in Fig.~\ref{fig:RC_scheme_a}(i).  Because of the RC fading memory, required for good performance \cite{Bertschinger2004}, bits earlier in the sequence are partially forgotten by the time the $n^{th}$ bit is injected into the reservoir. Also, information from one $n$-bit sequence spills into the next sequence.  Thus, the combination of serial-data-input and fading memory violates the sequence-order permutation symmetry. No adjustment of the RC hyperparameters can fully fix this symmetry mismatch and the problem becomes more pronounced as $n$ increases.
 
Furthermore, the parity-order symmetry and the Lorenz '63 system inversion are not respected by the standard RC commonly used in the reservoir computing community where $f(\mathbf{r})$=$\mathrm{tanh}(\mathbf{r})$, $g(\mathbf{r})$=$\mathbf{r}$, and $b$=0.  In this case, the RC possesses inversion symmetry $(\mathbf{u},\mathbf{r},\mathbf{v})\rightarrow-(\mathbf{u},\mathbf{r},\mathbf{v})$, which respects only the parity-order symmetry for $n$ odd, but not for $n$ even nor the Lorenz '63 system inversion symmetry. Thus, we expect poor performance for the latter two tasks. 
Prior work on RC has demonstrated high performance on the parity task for $n$ odd \cite{Bertschinger2004,Dasgupta2012,Snyder2013,Schumacher2013}, while related work where the RC does not fulfill the inversion symmetry rule has shown high performance for both odd and even $n$ \cite{Coulombe2017,Dion2018}. 
Prior work on the Lorenz '63 system prediction task has also shown an improvement in performance when the RC has a broken symmetry \cite{Pathak2017,Lu2017,Lu2018,Griffith2019,Herteux2020}. However, the literature does not explore the effects of symmetry breaking parameter changes or symmetry matching on reservoir performance.

We make changes to both the input and output layers to solve these problems and realize a symmetry-aware RC; no change to the reservoir is required.  To address the parity sequence-order permutation symmetry we make two changes to the input layer.  First, we use a tapped delay line for the input data as shown in Fig.~\ref{fig:RC_scheme_a}(ii), which converts the serial data into an $n$-bit parallel word. Serial-to-parallel conversion is a common method in high-speed electronics and hence can be achieved in hardware without loss of RC throughput.  
Here, the input is the $n$-dimensional vector
\begin{equation}
\mathbf{u}(t)={[u(t),u(t-T), \ldots u(t-[n-1]T)]}^\intercal , 
\end{equation}
\noindent  where $\,^\intercal$ indicates the transpose. 
Thus, all $n$ components are input into the reservoir simultaneously, while in the serial input scheme only a single bit is input during the time interval $T$. 
The second modification is to broadcast all $n$ components of the data vector to each node with identical weight determined by $W_{in}$. We also reset all reservoir nodes to zero after the time $T$ when a new sequence is input. These changes restore the sequence-order permutation symmetry.

The parity-order symmetry can be respected to some extent by changing the symmetry of $f$, $g$, or taking $b\neq$0.  However, changing the symmetry of $f$ affects the inhibitory versus excitatory aspect of the signals and hence can have a negative impact on RC performance. Similarly, it is difficult (or impossible, depending on $f$) to have a pure even or odd symmetry by adjusting $b$.  On the other hand, adjusting $g$ can provide symmetry matching by squaring a portion $\eta_r$ of nodes before the output multiplication so that 
\begin{equation}
    g(r_i)= 
\begin{cases}
    r_i^2,& \text{if } i\leq \eta_r N\\
    r_i,              & \text{if }  i>\eta_r N .
\end{cases}
\end{equation}

\noindent An optimization routine can be used to select $\eta_r$.  In Appendix \ref{appendix:b}, we compare all three approaches and demonstrate that adjusting only $g$ gives rise to a high-performing RC for the parity task.

To respect the Lorenz '63 system inversion symmetry, we make changes either in the input or in the output layer. For the first, we square the input signal so that the RC input-to-output relations are described by
\begin{equation}
\begin{aligned}
    u= [x,y]  \rightarrow & \,\,[x^2,y^2] \rightarrow \bf{r} \rightarrow \bf{v} \\
    -u=[-x,-y]  \rightarrow &\,\,[x^2,y^2] \rightarrow \bf{r} \rightarrow \bf{v},
\end{aligned}
\end{equation}

\noindent where both inputs $u=[x,y]$ and $-u=[-x,-y]$ lead to the same reservoir state $\bf{r}$ and consequently to the same output ${\bf{v}}(t) \widehat{=}  z(t)$, thus respecting the Lorenz 63' system symmetry.
For the later, the symmetry matching is obtained by adjusting $g$ just like in the case of the parity task for $n$ even. Here, when we set $\eta_r = 1$ the RC  input-to-output relations become 
\begin{equation}
\begin{aligned}
     u=[x,y]  \rightarrow  \bf{r} \rightarrow & \,\, \bf{r^2}\rightarrow \bf{v} \\
    -u=[-x,-y]  \rightarrow  -\bf{r} \rightarrow & \,\, \bf{r^2} \rightarrow \bf{v},
\end{aligned}
 \end{equation}
  \noindent where the inputs $u=[x,y]$ and $-u=[-x,-y]$ lead the reservoir to opposite states ${\bf{r}}$ and $-{\bf{r}}$, but the squared readout guarantees the same feature vector ${\bf{r^2}}$ and the symmetry matching between the RC and the learning system (here Lorenz 63' system). 
We use a serial input scheme for the inference task where, for each time, only the current value of $u$ is input into the reservoir, as shown in Fig. \ref{fig:RC_scheme_b}.  For all results presented below, we set $f(x)$=$\tanh(x)$ and $b$=0. 


\section{Results}
\label{sec:Results}
\subsection{Parity task}
We demonstrate that when both parity symmetries are taken into account, an RC can be designed to achieve zero error for the $P_n$ task using exponentially reduced neural network and training size in comparison to regular non symmetry-aware RCs.

\subsubsection{Non symmetry-aware RC}
As a baseline, we perform the parity task applied to a 1000-bit random test time-series data shown in the top panel of Fig.~\ref{fig:Inversion_Symmetry_Breaking_a} for $n$=6 using the common RC configuration of serial-data input with $\eta_r$=0 and $N$=100. The reservoir is trained using a different random binary time series with 1,000 bits and with optimized hyperparameters. Comparing the ground truth and RC-predicted parity in the bottom left panel of Fig.~\ref{fig:Inversion_Symmetry_Breaking_a}, we see that the RC performs poorly with a  bit error rate (BER) of 0.4 - essentially not much better than guessing.

\subsubsection{Respecting parity-order symmetry}
\begin{figure}[!t]
\centering
	\subfloat{%
	\includegraphics[width=\linewidth]{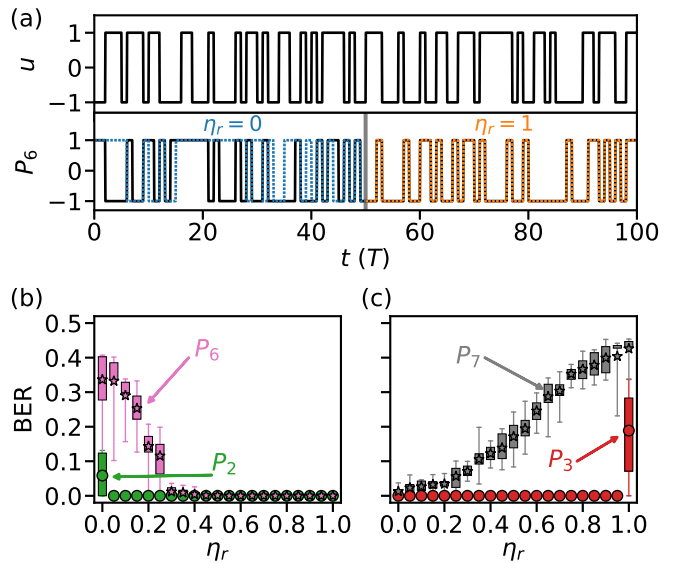}%
	\label{fig:Inversion_Symmetry_Breaking_a}%
	}
	\begin{minipage}[t]{0\textwidth}
	\subfloat{%
	\label{fig:Inversion_Symmetry_Breaking_b}%
	}
    \end{minipage}
    \begin{minipage}[t]{0\textwidth}
	\subfloat{%
	\label{fig:Inversion_Symmetry_Breaking_c}%
	}
	\end{minipage}
\caption{Parity task: RC performance as function of $\eta_r$. (a) Top: Segment of input testing signal $u$. Bottom: $P_6$ desired output (continuous black line) and the optimized RC output (dashed line) for $\eta_r$=0 (left)  and $\eta_r$=1 (right). The hyperparameters are $(T_0,\Delta T, \gamma, \rho_r, \sigma, \rho_{in})$=$(0.20T,0.45T,2.44T^{-1},1.26,0.72,0.30)$ and $(0.45T,0.40T,4.40T^{-1},1.58,0.99,0.93)$, respectively. (b) and (c) Mean BER of 10 optimized RC instances as a function of $\eta_r$. The vertical bars are limited by the $q_1$ and $q_3$ quartiles and the vertical lines by the minimum and maximum BER values.
} 
\label{fig:Inversion_Symmetry_Breaking}
\end{figure}

Next, we modify only the output layer by taking $\eta_r$=1 so that the parity-order symmetry is respected for this case when $n$ is even.  The reservoir is retrained and the hyperparameters re-optimized. Dramatically, the BER drops to zero as seen in the bottom right panel of Fig.~\ref{fig:Inversion_Symmetry_Breaking_a}, albeit for this fairly large reservoir. To our knowledge, there are no previous reports of obtaining zero-error for $P_6$ in the reservoir computing literature, demonstrating the importance of respecting the parity-order symmetry.

To explore this point further, we measure the BER as a function of $\eta_r$ as seen in Figs.~\ref{fig:Inversion_Symmetry_Breaking_b} and \ref{fig:Inversion_Symmetry_Breaking_c}. For each point, we optimize the hyperparameters for 10 different RCs. For $n$=2 or 3, the sequences are short enough that zero-error is obtained even when the symmetry is not fully satisfied ($\eta_r$ should be equal to 1 for $n$ even and 0 for $n$ odd to fully satisfy the parity-order symmetry).  However, for larger $n$, it is of greater importance to match this symmetry.  For $P_7$, the mean BER is $0.013$  with standard deviation of $0.009$ for $\eta_r$=0, demonstrating that satisfying the parity-order symmetry alone is not enough to obtain zero-error for this reservoir size.

We expect that the performance of the RC will improve as $N$ increases as is generally found in the RC literature. To explore the reservoir size required to obtain zero-error on the parity task, we set $\eta_r$ to respect the parity-order symmetry, instantiate 50 different RCs and optimize the hyperparameters for each. Figure \ref{fig:N_vs_n_vs_BER_a} shows the mean BER (color scale) for each $N$ and  $n$. Here, we stop increasing $N$ when all 50 RCs reach BER=0.  The width of the horizontal bars indicates the fraction of reservoirs with BER=0, where the minimum width for small $N$ indicating that no reservoir has zero-error.  The white star indicates the smallest $N$ for which at least one out of the 50 RCs obtains BER=0.  While we only go up to $n$=7 due to exponential increasing computational cost, the fitting (dashed line) shows an exponential scaling of $N$ to obtain BER=0 for these RCs that respect parity-order symmetry but use serial input. 
Here, the training and the testing data sets are composed by different 1000-bit random time series. We check these time series to make sure that all the $2^n$ different $n$-bit patterns are presented at least once to the RC in both training and testing phase.

\subsubsection{Respecting both parity-order and sequence-order permutation symmetries}

\begin{figure}[!t]
\centering
\subfloat{%
\includegraphics[width=\linewidth]{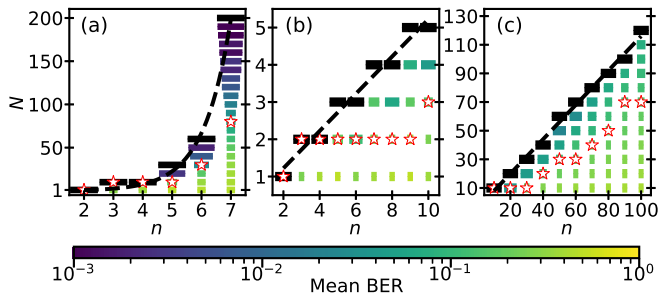}%
\label{fig:N_vs_n_vs_BER_a}%
}
\begin{minipage}[t]{0\textwidth}
\subfloat{%
\label{fig:N_vs_n_vs_BER_b}%
}
\end{minipage}
\begin{minipage}[t]{0\textwidth}
\subfloat{%
\label{fig:N_vs_n_vs_BER_c}%
}
\end{minipage}
\caption{Mean BER as function of $N$ and $n$. The dashed lines represent the fit of the network size scaling to obtain a mean BER=0 (black bars). 
(a) Only the parity-order symmetry is respected. The $y$-axis starts with $N$=1 and $N$=10, then $N$ is incremented by 10. The fit shows an exponential scaling with coefficient of determination $R^2$=0.994. (b) and (c) Both parity-order and sequence-order permutation symmetries are respected and the fit shows linear scaling $N\sim0.50n+0.22$ with $R^2$=0.96 for $n\le$10 and $N\sim1.2n-4.0$ with $R^2$=0.99 for $10\le n \le 100$, respectively.
}  
\label{fig:N_vs_n_vs_BER}
\end{figure}

We find a remarkable improvement in the RC performance when respecting both symmetries.  We use the parallel input scheme discussed above while simultaneously setting $\eta_r$ to satisfy the parity-order symmetry.  As seen in Fig. \ref{fig:N_vs_n_vs_BER_b}, we find that a reservoir with only $N\le$3 is enough to obtain BER=0 for up to $n$=7, an exponential reduction in $N$ in comparison to the serial-input case that does not respect the sequence-order permutation symmetry. To our knowledge, there are no previous results in the reservoir computing literature that completely solve the parity task using such small networks. Figure \ref{fig:N_vs_n_vs_BER_c} shows that $N$ continues linear scaling for $n$ up to 100. Past work using hand-crafted ANNs solved the parity task with a scaling of $N = \log_2(n+1)$ \cite{Hunter2012}, but full accuracy with such scaling rule was not obtained when training these ANNs architectures from initial random weights. Their success rate decreased with increasing n. 

As a final thought on using RCs for solving the parity task, we note that previous studies trained the RC with long random bit sequences.  Commonly, it is found that the performance increases with the length of the training set. We hypothesize that the reason the performance improves for longer random binary sequences is partly due to the fact that the RC is more likely to be presented with the entire set of unique sequences the longer the data set.

To quantify this point, we find that the expected number of $n$-bit-long sequences required in the training time series is given approximately by the coupon collector expression 
\begin{equation}
E(n)=1+\frac{2^n}{2^n-1}+\frac{2^n}{2^n-2}+...+\frac{2^n}{1}=2^nH_{2^n},
\label{eq:coupon_collector}
\end{equation}
\noindent where $H_M$ is the $M^{th}$ harmonic number \cite{Flajolet1990}.  Because the parity task involves a sliding window with $n$ bits being processed at a time, there is re-use of bits from one sequence to the next.  Accounting for this reuse, the training time series only need to contain, on average, $E(n)+n-1$ bits.  As an example, $E$=22 for $n$=3 so that we need to train the reservoir with a 24-bit-long random sequence on average.

For a fully symmetry-aware RC, each sequence in the set $L_n(l)$ is equivalent so the reservoir only needs to be trained on any one sequence in each set.  Furthermore, the NOT of a sequence in $L_n(l)$ (equivalent to $\mathbf{u}\rightarrow -\mathbf{u}$) is found in the set $L_n(n-l)$ and the parity-order symmetry ensures that the RC will give the correct result just by training on the sequence; that is, the NOT of the sequence is not needed.

To quantitatively predict the number of sequences required to train the reservoir based on this line of reasoning, we introduce the parameter $s$, which is the minimum number of 1's or -1's in a sequence.  Its maximum value $s_{max}$ is $n/2$ for $n$ even and $(n-1)/2$ for $n$ odd. With this notation, the number of $n$-bit-long sequences for training is $(s_{max} + 1)$.  Because of the sliding window and bit re-use mentioned above, the required training length is only $n+s_{max}$, an exponential reduction in comparison to the standard method of training a non-symmetry-aware RC.  A simple way to construct the training data set in this case is to make the first $n$ bits equal to -1 and the following $s_{max}$ bits equal to 1. We use this procedure on the RCs of Figs.~\ref{fig:N_vs_n_vs_BER_b} and \ref{fig:N_vs_n_vs_BER_c}, which greatly reduced the computation time to generate this plot in addition to the savings obtained by using a much smaller $N$.

\subsection{Inference task}

We demonstrate how the RC performance is improved for the inference task when the inversion symmetry in the Lorenz 63' system is taken into account. The RC can respect such symmetry either by changing $\eta_r$ at the output layer or by squaring the input signal, thus modifying the input layer. 
Similarly to our approach for solving the parity task, here we make changes only on either the input or output layer to match the input system symmetry and choose the reservoir randomly with no problem-based modifications. 
For performance comparison, we measure NRMSE  between the actual and the inferred variables. 

\subsubsection{Respecting symmetry by adjusting output layer}

\begin{figure}[!t]
\centering
\subfloat{%
\includegraphics[width=\linewidth]{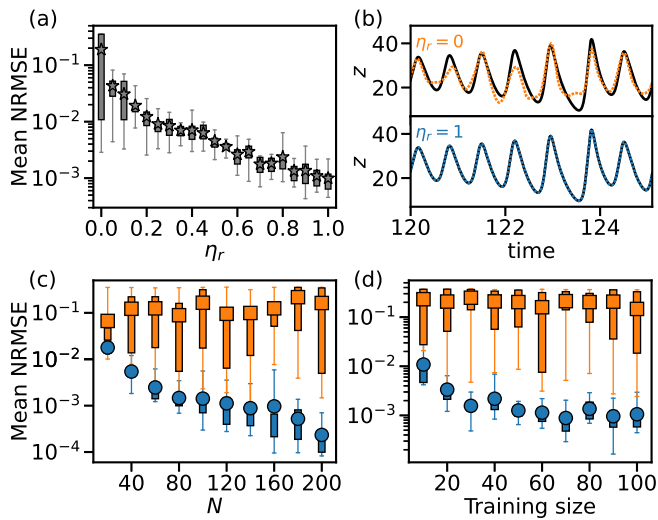}%
\label{fig:NRMSE_vs_eta_r}%
}
\begin{minipage}[t]{0\textwidth}
\subfloat{%
\label{fig:Inference_TimeSeries}%
}
\end{minipage}
\begin{minipage}[t]{0\textwidth}
\subfloat{%
\label{fig:3a}%
}
\end{minipage}
\begin{minipage}[t]{0\textwidth}
\subfloat{%
\label{fig:3b}%
}
\end{minipage}
\caption{Lorenz '63 chaotic system inference task with ${\bf{u}}=[x,y]$ as input: RC performance as function of $\eta_r$, $N$ and training size. (a)  Mean NRMSE of 10 optimized RCs as function of $\eta_r$ for $N=100$ and a training size of 100 units of time. (b) Actual (solid black line) and inferred (dashed line) $z$ for $N=100$. Top (orange): regular RC ($\eta_r = 0$) for optimal hyperparamters $(\gamma, \rho_r, \sigma, \rho_{in})$=$(16.09,1.12,0.001,0.53)$. Bottom (blue):  symmetry-aware RC ($\eta_r=1$)  for optimal hyperparamters $(\gamma, \rho_r, \sigma, \rho_{in})$=$(14.29,0.87,0.06,0.32)$.  (c) Mean NRMSE of 10 optimized RCs as function of $N$ for $\eta_r=0$ (orange squares) and for $\eta_r=1$ (blue circles) with a fixed training size of 100 units of time. (d) NRMSE of 10 optimized RCs as function of  training size for $\eta_r=0$ (orange squares) and for $\eta_r=1$ (blue circles) with a fixed reservoir size $N=100$. Unless declared otherwise, the training and testing data sizes are 100 units of time each with a fixed sample time of $0.005$. The vertical bars are limited by the $q_1$ and $q_3$ quartiles and the vertical lines by the minimum and maximum NRMSE values.
}  
\label{fig:Inference_NRMSE}
\end{figure}

First, we consider only adjustments in $\eta_r$ and use ${\bf{u}}=[x,y]$ as input.  Figure \ref{fig:NRMSE_vs_eta_r} shows the mean NRMSE of 10 different optimized RCs as function of $\eta_r$.  The hyperparameters were optimized for each RC and the reservoir and training sizes were kept fixed to $N=100$ and 100 time units, respectively. The error decreases with the increasing of $\eta_r$ towards the symmetry matching parameter value ($\eta_r=1$). Segments of the actual variable $z$ and its inference done by a given RC instance are shown in Fig. \ref{fig:Inference_TimeSeries}. The RC performs poorly when $\eta_r=0$ (regular non symmetry-aware RC) resulting in a  NRMSE $=0.14166$. When the Lorenz '63 system symmetry is respected by setting  $\eta_r=1$ (symmetry-aware RC), the NRMSE drops to $0.00046$, improving the performance by three orders-of-magnitude. 

It is commonly found in the RC literature that the RC performance improves as $N$ and the training size increase. To observe how the mean NRMSE depends on the reservoir size we fixed the training size to 100 time units while $N$ is varied. Figure \ref{fig:3a} shows the performance depence on $N$ for both a regular non symmetry-aware RC with $\eta_r=0$ (orange squares) and a symmetry-aware RC whose symmetry is matched in the output layer by setting $\eta_r=1$ (blue circles). For the first, the RC performs poorly with an NRMSE around 0.1 independent of the network size. 

On the other hand, when the symmetry is respected, the performance is improved as $N$ increases. For $N=200$ the mean NRMSE is improved by three orders-of-magnitude in comparison to the regular RC. This indicates that our reservoir implementation presents a high-dimensional state space large enough to provide a good computational capacity to solve this task. Thus, we conclude that the poor performance of the standard RC for the inference task is mainly related to symmetry mismatch between the RC and the Lorenz 63' system rather than lack of either computational capacity or parameter optimization (all hyperparameters are optimized for each RC instance).

The dependence of the mean NRMSE on the training size is shown in Fig. \ref{fig:3b}. We keep $N=100$ fixed and vary the training size. For the case where symmetry is matched, the performance improves by one order of magnitude when increasing the training data size. Here, we highlight the generalization capacity of the symmetry aware RC. Even though only a small part of the chaotic attractor is presented to the reservoir during a small training period, the symmetry aware RC demonstrates its capacity to generalize by correctly inferring the unknown variable with NRMSEs as small as $10^{-3}$ for regions of the attractor never seen during training. Even for training data sets as small as 30 time units, the symmetry aware RC performs with NRMSE two orders-of-magnitude better than a regular RC, which in itself is known to be less data hungry than other methods like deep neural networks for two main reasons. First, RCs have less trainable parameters once only the output layer is trained. Second, the training process is usually a simple linear regression instead of a nonlinear optimization. Recently, it was shown that a Next Generation Reservoir Computer (NG-RC) can perform predictions and inference tasks better than regular RCs with even less training data \cite{NG-RC}.

\subsubsection{Respecting symmetry by adjusting input layer}

\begin{figure}[t]
\centering
\subfloat{%
\includegraphics[width=\linewidth]{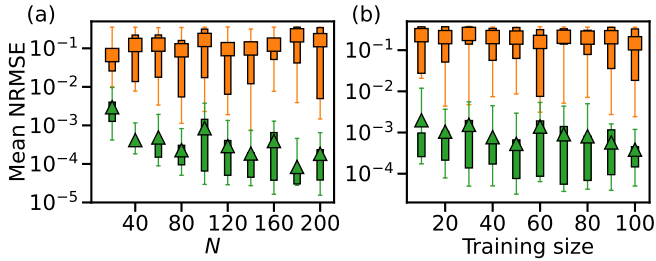}%
\label{fig:5a}%
}
\begin{minipage}[t]{0\textwidth}
\subfloat{%
\label{fig:5b}%
}
\end{minipage}
\caption{Comparison between mean NRMSE of 10 optimized regular RCs (orange squares) and 10 optimized symmetry-aware RCs which have the symmetry matched to the input data by squaring the input ${\bf{u}}=[x^2,y^2]$ (green triangles). (a) mean NRMSE as function of $N$ for a fixed training size of 100 time units. (b) mean NRMSE as function of the training size for a fixed reservoir size of $N=100$. In both cases we set $\eta_r=0$ and optimize the hyperparameters for each RC instance. The vertical bars are limited by the $q_1$ and $q_3$ quartiles and the vertical lines by the minimum and maximum NRMSE values.} 
\label{fig:5}
\end{figure}

Finally, we investigate the RC performance when the symmetry is matched by adjusting only the input layer. For that, we square the input data so that ${\bf{u}}=[x^2,y^2]$ and keep $\eta_r=0$, $\textit{i.e.}$, we do not adjust the output layer symmetry breaking parameter. Figure \ref{fig:5} shows the mean NRMSE as function of $N$ and the training size for this case. The green triangles symbols are for the symmetry-aware RC with input squared and, for a better comparison, we repeat the plots of the NRMSE for the standard RC (orange squares) from Fig. \ref{fig:Inference_NRMSE}.  
Similarly to the case where the symmetry is matched in the output layer, here the symmetry-aware RC presents a mean NRMSE up to three orders-of-magnitude lower than the regular RC as shown in Fig. \ref{fig:5a}.

As a last thought, we highlight that, for both methods of symmetry matching presented in this work for the inference task, the performance does not improve for training sizes longer than 50 time units as shown in Figs. \ref{fig:3b} and \ref{fig:5b}. 
An increase in the training dataset is expected to improve the performance of the RC, as a larger region of the attractor is presented to the network. However, further studies are need to investigate the reason why the performance improvement saturates for such small training data size.

\section{Conclusion}
\label{sec:Conclusion}
Our work highlights the importance of matching the symmetry of an RC to the symmetry of the data being processed and the fact that these symmetries can be satisfied by only making changes to the input and output layers of the RC. The parallel input scheme and the input squaring procedure are used to match specific symmetries in the parity task and in the inference tasks, respectively.  On the other hand, the output layer symmetry breaking parameter $\eta_r$ is introduced and tuned until the RC runs best, meaning that we can discover whether we need to match or break the RC symmetry according to the input data. Both methods are valuable: modify the RC to account for symmetries we know exist, and then try to introduce parameters for symmetries we suspect exist.   

Of note is the observation that a symmetry-aware RC has vastly improved performance.  For the parity task, traditionally considered a hard ML problem, we obtain an exponential reduction in the network and training set sizes needed to obtain zero-error. For the chaotic system inference task we obtain a performance three orders-of-magnitude better than regular RCs. In principle, the symmetry considerations we have used to achieve drastic improvement in performance for reservoir computing can be applied to other neuromorphic and machine learning approaches, such as ANNs. Future research is required to determine if similar performance improvements can be found in these methodologies when symmetry is a design consideration.


\begin{acknowledgments}
W.A.S.B. and D.J.G. gratefully acknowledge the financial support of Raytheon BBN Technologies through project \#60150.
\end{acknowledgments}


\appendix

\section{Hyper-parameters Optimization}
\label{appendix:a}

\begin{table}[b]
\caption{\label{tab:parameters}%
Hyper-parameter space scanned by the Bayesian optimizer.  \\}
\begin{ruledtabular}
\begin{tabular}{cccc}
\multicolumn{1}{p{1cm}}{\centering  Hyper-parameter}& 
\multicolumn{1}{p{2cm}}{\centering Parity task \\ serial input} &
\multicolumn{1}{p{2cm}}{\centering Parity task \\ parallel input} &
\multicolumn{1}{p{2cm}}{\centering Lorenz '63 \\ inference task } \\
\colrule \\
$T_0$ [$T$] & 0-0.5 & 0-1 & -  \\ 
$\Delta T$ [$T$] & 0.05-0.5  & 0.05-1& -  \\ 
$\gamma$ & 0.1-5.0& 0.1-10.0 & 0.01-20.0 \\ 
$\rho_r$  & 0.1-2.0 & 0.1-10.0&0.001-5.0  \\ 
$\rho_{in}$ & 0.1-1.0& 0.1-1.0 & 0.001-1.0  \\ 
$\sigma$   & 0.1-1.0 & 0.1-1.0 & 0.01-1.0  
\end{tabular}
\end{ruledtabular}
\end{table}

We use a Gaussian-Process-based Bayesian optimizer available in the {\it {skopt}} {\it {python}} module to find the optimal hyperparameters ($T_0$,$\Delta T$,$\gamma$,$\rho_r$,$\sigma$,$\rho_{in}$).
For the parity task, we keep $k=10$ ($k=N$ for $N<10$) for the serial and $k=1$ for parallel input schemes. We integrate the reservoir with a simple  Euler  algorithm  with  time  step $dt$= 0.01$T$ for the serial input scheme and save the reservoir state every 5 integration steps. For the parallel input scheme, we use $dt$= 0.001$T$ to have an integration time step 100 times smaller than the characteristic decay time of the nodes ($1/\gamma$) which can be as small as 0.1 in this case (see Table \ref{tab:parameters}). For the parallel input scheme, we save the reservoir state every 50 integration steps. For the inference task,  we keep $k=5$ and integrate the Lorenz '63 system and reservoir equations with integration steps $0.0001$ and $0.005$, respectively.


\begin{figure}[t]
\centering
\includegraphics[width=\linewidth]{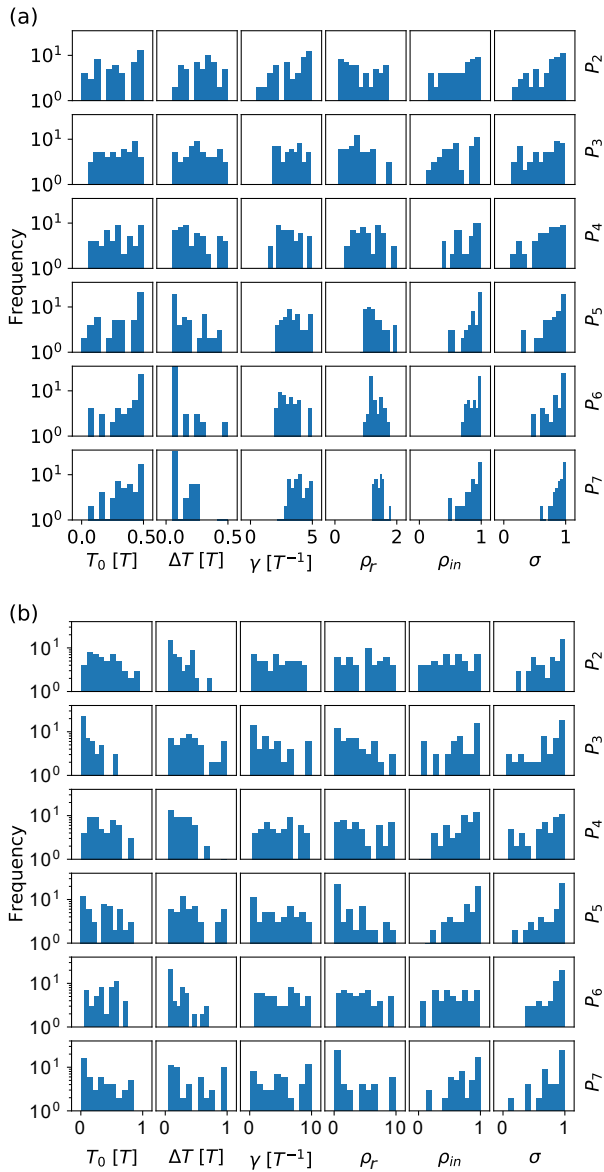}\\
\caption{Optimal parameters distribution of the 50 RCs that have BER = 0 for $2 \le n \leq 7$ for (a) the serial input scheme where only the parity-order symmetry is respected and (b) the parallel input scheme where both parity-order and sequence-order permutation symmetries are respected.} 
\label{fig:OptimalParameters50RCs}
\end{figure}


Table \ref{tab:parameters} shows the scanned range for each hyper-parameter for each task. The optimal hyper-parameters may change for different RC topologies, {\textit{i.e.}}, for different $W_r$ and $W_{in}$, which are chosen before optimization. Thus, most of the hyper-parameters do not have a preferred optimal value. As an example of such diversity, Fig.  \ref{fig:OptimalParameters50RCs} shows the optimal hyper-parameters distribution of the 50 RCs that have BER = 0 for the parity task in Fig. \ref{fig:N_vs_n_vs_BER_a} and Fig. \ref{fig:N_vs_n_vs_BER_b}. The all set of optimal hyper-parameters for the parity task and for the Lorenz '63 system inference is available upon reasonable request.

\section{RC Symmetry Breaking Parameters}
\label{appendix:b}

The RC inversion symmetry can be adjusted by three different ways:
\noindent
\begin{itemize}	[labelsep=*,leftmargin=1pc]
\item Changing the symmetry of $f$: we use $f=\tanh^2$ as the nonlinearity for a portion $\eta_f$ of the nodes. 
\item Changing the symmetry of $g$: we square ${\bf{r}}(t)$ for the portion $\eta_r$ of nodes just before output matrix multiplication.
\item Adding a bias $b$: we introduce a bias $b\neq 0$ in the argument of $f$.
\end{itemize}

\begin{figure}[h]
\centering
\includegraphics[width=0.75\linewidth]{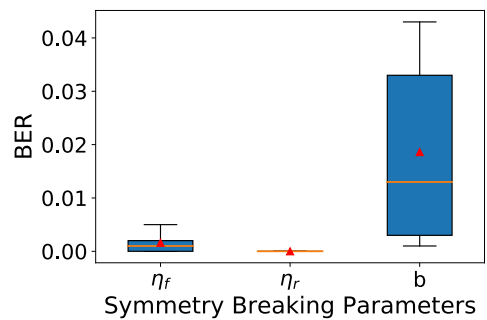}
\caption{$P_6$ classification BER for $\eta_f$, $\eta_r$ and $b$ as symmetry breaking parameter. The box plot represents a set of 5 optimized RC instances. The mean BER  is represented by the red triangles, the blue box is limited by the $q_1$  and $q_3$ quartiles, the orange horizontal line stands for the median and the vertical lines are limited by the minimum and maximum BERs among the 5 instances. } 
\label{fig:SymmBreakParameters}
\end{figure}

Figure \ref{fig:SymmBreakParameters} shows a box plot for the $P_6$ classification BER for when the RC has its symmetry adjusted separately by $\eta_f$, $\eta_r$ and $b$. When one of these three parameters is adjusted, the other two are set to zero. For each case, 5 different RC instances are optimized. The mean BER is represented by the red triangles.

We find that the best RC performance (mean BER = 0) is obtained when we adjust $\eta_r$. For this case, the symmetry is broken at the output layer and all the network nodes can take on negative or positive values. This does not happen when we break the symmetry by adjusting $\eta_f$. In that case, a portion of nodes has its state set to be always positive due to its nonlinearity $f=\tanh^2$. These nodes are always excitatory to the rest of the network. This may limit the network inhibitory behavior and decrease the network computational capacity. Adjusting the bias is the worst of the three symmetry breaking procedures. The high mean BER for $P_6$ classification in comparison to the other two parameters is explained by the inability of the RC of having an even function whenever there is a bias inside the nonlinear function $f=\tanh$. Also, the bias can saturate the node state making it less sensitive to external and internal inputs.


\bibliographystyle{apsrev4-1}

%

\end{document}